\documentclass[letterpaper]{article} 
\usepackage{aaai20}  
\usepackage{times}  
\usepackage{helvet} 
\usepackage{courier}  
\usepackage[hyphens]{url}  
\usepackage{graphicx} 
\usepackage{graphicx}  
\usepackage{xspace}
\usepackage{amsmath}
\usepackage{amsfonts}
\usepackage{quoting}
\usepackage{booktabs}
\usepackage{multirow}

\newcommand{\paratitle}[1]{\noindent \textbf{#1}}

\newcommand{\ie}{\emph{i.e.,}\xspace}
\newcommand{\eg}{\emph{e.g.,}\xspace}

\newcommand{\etc}{\emph{etc.}\xspace}

\newcommand{\citenoun}[1]{\citet{#1}}
\newcommand{\citet}[1]{\citeauthor{#1} \shortcite{#1}}

\title{CASE: Context-Aware Semantic Expansion}
\author{Jialong Han\textsuperscript{\rm 1}\thanks{Work done when Jialong Han was with Tencent AI Lab.}, Aixin Sun\textsuperscript{\rm 2}, Haisong Zhang\textsuperscript{\rm 3}, Chenliang Li\textsuperscript{\rm 4}, Shuming Shi\textsuperscript{\rm 3}\\
\textsuperscript{\rm 1}Amazon, USA, \textsuperscript{\rm 2}Nanyang Technological University, Singapore, \textsuperscript{\rm 3}Tencent AI Lab, China, \textsuperscript{\rm 4}Wuhan University, China\\
\textsuperscript{\rm 1}jialonghan@gmail.com, \textsuperscript{\rm 2}axsun@ntu.edu.sg, \textsuperscript{\rm 3}\{hansonzhang, shumingshi\}@tencent.com, \textsuperscript{\rm 4}cllee@whu.edu.cn
}

\makeatletter
 \def\@biblabel#1{}
 \makeatother

\begin{document}

\maketitle

\begin{abstract}
In this paper, we define and study a new task called \emph{Context-Aware Semantic Expansion} (CASE).
Given a \emph{seed term} in a sentential context, we aim to suggest other terms that well fit the context as the seed.
CASE has many interesting applications such as query suggestion, computer-assisted writing, and word sense disambiguation, to name a few.
Previous explorations, if any, only involve some similar tasks, and all require human annotations for evaluation.
In this study, we demonstrate that annotations for this task can be harvested \underline{at scale} from existing corpora, in a fully automatic manner.
On a dataset of 1.8 million sentences thus derived, we propose a network architecture that encodes the context and seed term separately before suggesting alternative terms.
The context encoder in this architecture can be easily extended by incorporating seed-aware attention.
Our experiments demonstrate that competitive results are achieved with appropriate choices of context encoder and attention scoring function.
\end{abstract}

\section{Introduction}\label{sec:intro}

Have you ever googled ``\emph{Lionel Messi championships}'', browsed the results, and wanted more soccer stars with comparable championships?
Have you ever wanted to know types of nutrients rich in barley grass, but were only able to remember amino acid?
In this paper, we study \emph{context-aware semantic expansion} (or CASE for short).
In CASE, user provides a \emph{seed term} wrapped in a sentential \emph{context} as in Figure~\ref{fig:example}.
The system returns a list of \emph{expansion terms}, each of which is a valid substitute for the seed, \ie the substitution is supported by some sentence in a (testing) corpus.
This task is not easy due to the large number of potential expansions, as well as the necessity of modeling their interactions with both the context and the seed.
Despite the challenge, the task is of practical importance and benefits many applications. We list a few examples here.

\paratitle{Query suggestion}~\cite{wen2001clustering}\textbf{.} In the aforementioned query ``\emph{Lionel Messi championships}'', keywords ``\emph{Lionel Messi}'' can be a seed term to expand, and a CASE system may suggest related entities, \eg ``\emph{Christiano Ronaldo}'', as expansion terms. Those terms may be used to suggest queries like ``\emph{Christiano Ronaldo championships}''.

\paratitle{Computer-assisted writing}~\cite{liu2011computer}\textbf{.} For casual or academic writing, exemplifications often help to explain and convince.
It is desirable to suggest contextually appropriate alternative words when an author can think of only one.

\paratitle{Other NLP tasks.} CASE can potentially enhance natural language processing (NLP) tasks.
For example, in word sense disambiguation~\cite{navigli2009word}, an ambiguous word like ``apple'' can be first expanded \emph{w.r.t.}\ its context.
The suggested context-aware terms (\eg fruits or companies)  provide cues for the disambiguation task.

\begin{figure}
  \begin{quoting}[leftmargin=0.8cm]
  \textbf{Seed in context}: ``Young barley grass is high in \underline{amino acid}.''\\
  \textbf{Expansion terms}: vitamin, antioxidant, enzyme, mineral, chlorophyll, \dots
  \end{quoting}
  \caption{A seed term ``amino acid'' in context. Here, ``fat'' is an invalid substitute. It is irrelevant to barley grass and tends not to be supported by general corpora.}\label{fig:example}
\end{figure}

\subsection{Comparison with Related Tasks}

Despite its significance, explorations on CASE remain limited.
\emph{Lexical substitution}~\cite{mccarthy2007semeval} is the most similar task to CASE.
Given a word in a sentential context, \eg ``the \underline{bright} girl is reading a book'', lexical substitution predicts synonyms fitting the context, \eg ``wise'' or ``clever'' rather than ``shining''.
The synonym candidates generally come from high-quality but relatively small dictionaries like WordNet~\cite{fellbaum1998wordnet}.
Compared with lexical substitution, candidate expansion terms of CASE, \eg entity names, are not required to be aliases of the seeds but could be far more in number and less organized.


Besides lexical substitution, another task similar to CASE is \emph{set expansion}~\cite{tong2008system,wang2007language,he2011seisa,chen2016long,shen2017setexpan,shi2010corpus}.
It is to expand a few seeds (\eg amino acid and vitamin) to more terms in the same semantic class (\ie nutrition).
However, set expansion does not involve possible textual contexts with the seeds.
This may cause ``fat'' to appear in the results of Figure~\ref{fig:example}, which is irrelevant to barley grass.

While the above two tasks differ considerably from CASE by task definition, we further note that their model tuning and evaluation require manual annotations, which are hard to collect at scale.
Fortunately, CASE benefits from large-scale \emph{natural} annotations, as described below.

\subsection{Dataset and Formal Task Definition}\label{sec:task}

A first step toward CASE with today's deep learning machinery is to accumulate large-scale annotations.
For this task, an ideal piece of annotation would be different terms appearing separately in identical contexts.
While this form of annotations is hard to obtain manually and rare in natural corpora, we note that people often list examples, which effectively serve as natural annotations.

In a general corpus, lists of examples usually follow Hearst patterns~\cite{hearst1992automatic,snow2005learning}, \eg ``\textsf{h} such as \textsf{t\textsubscript{1}}, \textsf{t\textsubscript{2}}, \dots'', ``\textsf{t\textsubscript{1}}, \textsf{t\textsubscript{2}}, \dots, and other \textsf{h}'', \etc
Here \textsf{h} denotes a \emph{hypernym}, and \{\textsf{t\textsubscript{i}}\} are \emph{hyponyms}.
We note that, in the context, if all hyponyms other than one is removed, the sentence is still ``correct'' in the sense of the corpus.

By post-processing a web-scale corpus (detailed in experiments), we derive a collection of 1.8 million naturally annotated sentences.
All of them are of the form $\langle C,T\rangle$ as below.

\begin{tabular}{p{3in}}\\
  \textbf{Context $C$}: ``Young barley grass is high in \underline{\ \ \ \ } and other phyto-nutrients.''\\
  \textbf{Terms $T$}: \{vitamin, antioxidant, enzyme, mineral, amino acid, chlorophyll\}\\
  \vspace{1mm}
\end{tabular}
\noindent Here $C$ is the sentential context with a \emph{placeholder} ``\underline{\ \ \ \ }''.
$T=\{t_i\}$ are hyponyms appearing at the placeholder in Hearst patterns.
The CASE task is to use a \emph{seed} term $s\in T$ and the context $C$ to recover the remaining terms $T\setminus\{s\}$.


Taking advantage of the large dataset derived, we propose a neural network architecture with attention mechanism~\cite{bahdanau2014neural} to learn supervised expansion models.
Readers may notice that, due to the use of Hearst patterns, the context $C$ above has an additional hypernym ``phyto-nutrient'' compared with Figure~\ref{fig:example}.
In experiments, in addition to comparisons among solutions, we will also study the impact of this gap.

To summarize, our contributions are:
\begin{itemize}
  \item We define and study a novel task, \ie CASE, which supports many interesting and important applications.
  \item We identify an easy yet effective method to collect natural annotations.
  \item We propose a neural network architecture for CASE, and further enhance it with the attention mechanism. On millions of naturally annotated sentences, we experimentally verify the superiority of our model.
\end{itemize}

\section{Model Overview}\label{sec:model}

Given the inherent variability of natural language and the sufficient annotations, we tackle CASE by a supervised neural-network-based approach.

\begin{figure}[t]
  \centering
    \includegraphics[width=\columnwidth]{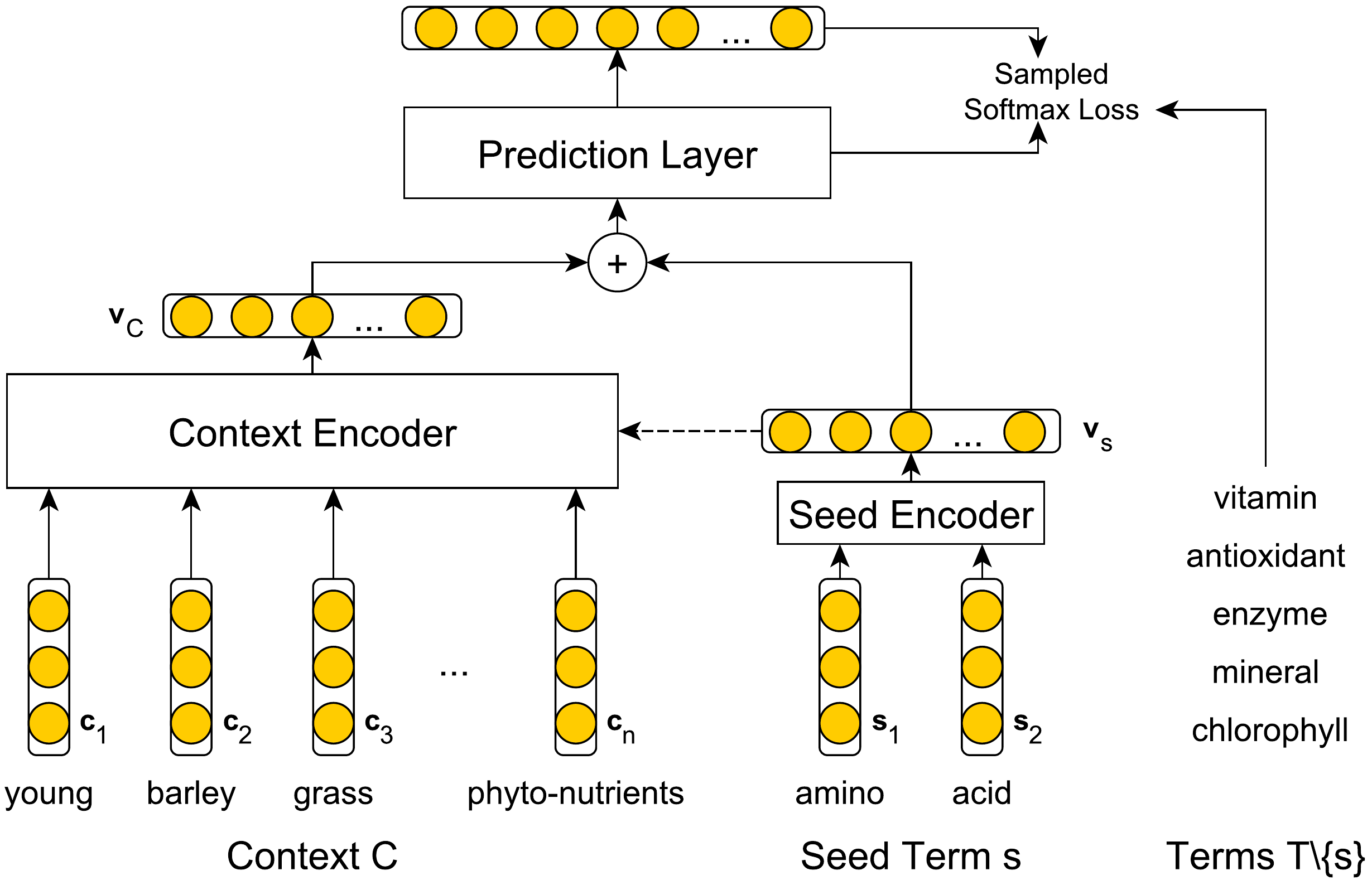}
  \caption{The network architecture of CASE.}
  \label{fig:network}
\end{figure}

Our network to model $P(.|s,C)$ is shown in Figure~\ref{fig:network}.
The network consists of three parts: a \emph{context encoder}, a \emph{seed encoder}, and a \emph{prediction layer}.
Given a seed $s$ in a context $C$, the network encodes them into two vectors $\mathbf{v}_s$ and $\mathbf{v}_C$ with the seed and context encoders, respectively.
The two vectors are then concatenated as input to the prediction layer to predict potential expansion terms.

On training sentences $\mathcal{T}$, we aim to optimize:
\begin{equation}\label{eq:obj}
\max \sum_{\langle C,T\rangle \in \mathcal{T}} \sum_{s\in T} \log P(T\setminus\{s\}|s,C)
\end{equation}
Note that, a sentence $\langle C,T\rangle$ is regarded as $|T|$ training samples.
Each sample treats one term as the seed, and predicts the other terms within the context.
In the remainder of this section, we briefly describe each of the three components.

\subsection{Encoding Sentential Contexts}

Given one or two seed terms, the traditional set expansion task simply finds other terms in the same semantic class.
However,  the sentential context $C$ may contain additional descriptions or restrictions, thus narrowing down the scope of the listed terms.
Therefore, it is vital to appropriately model $C$ to capture its underlying information in CASE.

In Figure~\ref{fig:network}, we employ the context encoder component to encode a variable-length context $C$ into a fixed-length vector $\mathbf{v}_C$.
There are various off-the-shelf neural models to encode sentences or sentential contexts.
On the one hand, by treating $C$ as a bag or a sequence of words, conventional sentence encoders may be applied, \eg Neural Bag-of-Words (\textsc{NBoW})~\cite{kalchbrenner2014convolutional}, \textsc{RNN}~\cite{pearlmutter1989learning}, and \textsc{CNN}~\cite{lecun1989backpropagation}.
On the other hand, there are also techniques that explicitly model placeholders, \eg CNN with positional features~\cite{zeng2014relation} and \textsc{context2vec}~\cite{melamud2016context2vec}.
In this paper, we mainly investigate \textsc{NBoW}-based and \textsc{RNN}-based encoders.
We also involve other encoders for comparison, \eg \textsc{CNN}-based and placeholder-aware encoders.


\paratitle{Neural Bag-of-Words Encoder.}
Given words $\{c_i\}_{i=1}^n$ in a context $C$, an \textsc{NBoW} encoder looks up their vectors $\mathbf{c}_i\in \mathbb{R}^d$ in an embedding matrix, and average the vectors as $\mathbf{v}_C=\frac{1}{n} \sum_{i=1}^{n} \mathbf{c}_i$.
The word embedding matrix is initialized with embeddings pre-trained on the original sentences in training set, and is updated during training.
Due to its simplicity, \textsc{NBoW} is efficient to train.
However, it ignores the order of context words. 


\paratitle{RNN- and CNN-Based Encoders.}
To study the impact of word order on context encoding, we consider \textsc{RNN}-based encoders as alternatives to \textsc{NBoW}.
\textsc{RNN}s take a sequence of context word vectors $(\mathbf{c}_1,\mathbf{c}_2,\dots,\mathbf{c}_n)$, and iteratively encodes information before each position $i$ as a sequence of \emph{hidden vectors} $(\mathbf{h}_1,\mathbf{h}_2,\dots,\mathbf{h}_n)$, $\mathbf{h}_i \in \mathbb{R}^d$.
Following~\citet{wang2016attention}, we take the last hidden vector $\mathbf{h}_n$ as the context vector $\mathbf{v}_C$:
\begin{align}
\mathbf{h}_i&=\text{RNN}(\mathbf{c}_i, \mathbf{h}_{i-1})\text{,}\quad i=1\dots n\text{,}\\
\mathbf{v}_C&=\mathbf{h}_n\text{.}
\end{align}

Besides the vanilla version of \textsc{RNN}, other \textsc{RNN} variations like \textsc{LSTM}~\cite{hochreiter1997long}, \textsc{GRU}~\cite{chung2014empirical}, and bi-directional \textsc{LSTM} (\textsc{BiLSTM})~\cite{graves2005framewise} have proven effective in various NLP tasks.
In our experiments, we compare all these RNN variations.

Other than the \textsc{NBoW}- and \textsc{RNN}-based encoders described above, \textsc{CNN}s~\cite{lecun1989backpropagation} have also been used as sentence encoders~\cite{kalchbrenner2014convolutional,kim2014convolutional,hu2014convolutional}.
Specifically, we perform the convolution operation on the input vector sequence $(\mathbf{c}_1,\mathbf{c}_2,\dots,\mathbf{c}_n)$, and apply max-pooling to get the context representation $\mathbf{v}_C$.


\paratitle{Position-Aware Encoders.}
All above encoders ignore the the position of the placeholder, \ie where the seed term appears.
For CASE, one may hypothesize that words at different distances to the placeholder contributes differently to $\mathbf{v}_C$.
\citenoun{zeng2014relation} propose \textsc{CNN} with positional features (\textsc{CNN+PF}) as a counterpart for \textsc{CNN}.
Each context word vector $\mathbf{c}_i$ fed into \textsc{CNN} is concatenated with a \emph{position vector} $\mathbf{p}_i$ that models its distance to the placeholder.
The positional vectors are treated as parameters and updated during training.
In \textsc{context2vec}~\cite{melamud2016context2vec}, two \textsc{LSTM}s are used to encode the left and right contexts of placeholders, respectively.
The output are concatenated as the final context representation $\mathbf{v}_C$.
We implement and compare it with \textsc{BiLSTM} as a counterpart.

\subsection{Encoding the Seed Term}

Due to its short length, we simply adopt the same \textsc{NBoW} to encode seed term, for it is less prone to overfitting~\cite{shimaoka2017neural}.
Given words $\{s_i\}_{i=1}^m$ of a seed term $s$, we obtain $\mathbf{v}_s$ by $\mathbf{v}_s=\frac{1}{m} \sum_{i=1}^{m} \mathbf{s}_i$.
Because of their different role, seed word embeddings $\mathbf{s}_i\in \mathbb{R}^d$ are from another embedding matrix, but are initialized and updated in the same manner with context word embeddings.

\subsection{Predicting Expansion Terms}

After encoding the seed and the context into $\mathbf{v}_s$ and $\mathbf{v}_C$, respectively, we feed their concatenation $\mathbf{x}=\mathbf{v}_s\oplus \mathbf{v}_C$ to the \emph{prediction layer} for expansion terms.
We treat the prediction as a classification problem, and each candidate term as a classification label.
Given a sufficiently large $\mathcal{T}$, we consider all terms appearing in Hearst pattern lists in $\mathcal{T}$ as candidates, and constitute the label set $L$ by pooling them, \ie $L=\cup_{\langle C,T\rangle \in \mathcal{T}}T$.
The prediction layer is then instantiated by a fully connected layer (with bias) followed by a softmax layer over $L$.
The probability of a term $t$ is then
\begin{equation}
P(t|s,C)=\frac{\exp(\mathbf{w}_t^\top\mathbf{x}+b_t)}{\sum_{t'\in L}\exp(\mathbf{w}_{t'}^\top\mathbf{x}+b_{t'})}\quad\text{.}\label{eq:soft_max}\\
\end{equation}
Here $\{\textbf{w}_t,b_t\}_{t\in L}$ are weight and bias parameters of the fully connected layer.

Note that we simultaneously predict multiple terms, \ie $T\setminus\{s\}$, so the classification is essentially multi-labeled.
Moreover, the softmax layer introduces summed exponentials on the denominator of $P(t|s,C)$.
This makes training inefficient on a large $L$ (over 180k on our dataset).
To relieve both issues, we use a multi-label implementation\footnote{\url{https://www.tensorflow.org/api_docs/python/tf/nn/sampled_softmax_loss}} of sampled softmax loss~\cite{jean2015using}.
That is, a much smaller \emph{candidate set} from $L$ is sampled to approximate gradients related to $L$.

\section{Incorporating Attention on Contexts}


So far, we have detailed various encoders for context $C$.
They all essentially aggregate the information in every word with or without position information in $C$.
Given potentially long input $C$ and the fixed output dimension, it is vital for encoders to capture the most useful information into $\mathbf{v}_C$.

Recent studies~\cite{bahdanau2014neural,luong2015effective,wang2016attention,shimaoka2017neural} suggest that attention-based encoders can focus on more important parts of sentences, thus achieving better representations.
In this section, we explore approaches to incorporate attention into the context encoders.
Based on whether they exploit information in the seed term, we categorize them as \emph{seed-oblivious} or \emph{seed-aware}.

\subsection{Seed-Oblivious Attention}

By seed-oblivious attention, we aim to model the importance of different words or positions in a sentential context.
Following conventional approaches~\cite{bahdanau2014neural}, we use a feed-forward network to estimate the importance of each word or position.
For the \textsc{NBoW} encoder, the importance score of word $i$ is defined by
\begin{align}\label{eq:seed-oblivious}
f(i)&=\mathbf{w}_a^\top \tanh(\mathbf{W}_a \mathbf{c}_i)\text{.}
\end{align}
Here $\mathbf{w}_a\in \mathbb{R}^{d'}$ and $\mathbf{W}_a\in \mathbb{R}^{d'\times d}$ are parameters of the feed-forward network.
The score $f(i)$ is then fed through a softmax layer and used as weights to combine the word vectors $\mathbf{c}_i$:
\begin{align}
\alpha_i&=\frac{\exp(f(i))}{\sum_{i'}\exp(f(i'))}\text{,}\label{eq:softmax}\\
\mathbf{v}_C&=\sum_{i=1}^{n} \alpha_i \mathbf{c}_i\text{.}\label{eq:attn_combine}
\end{align}
For \textsc{RNN}, attention is applied in a similar manner, except that $\mathbf{c}_i$ is substituted by hidden vector $\mathbf{h}_i$.


\subsection{Seed-Aware Attention}

So far, we have discussed various context encoders and an attention-based improvement. For them, the seed does not contribute to context encoding.
However, a seed like ``amino acid'' may conversely indicate informative words or parts in the context, \eg ``barley'' and ``grass'', to further narrow down the semantic scope of expansion.
Following this observation, we propose involving the seed vector $\mathbf{v}_s$ to compute a seed-aware importance score $f(s,i)$ instead of $f(i)$.
Inspired by~\citenoun{luong2015effective}, we consider the following instantiations of seed-aware attention.

\paratitle{\textsc{dot}}\quad In this variant, we estimate the word importance with the inner product of the seed vector $\mathbf{v}_s$ and each word vector $\mathbf{c}_i$.
Formally, the score is
\begin{align}
f(s,i)=\mathbf{v}_s^{\top}\mathbf{c}_i\text{.}
\end{align}

\paratitle{\textsc{concat}}\quad Instead of directly taking the inner product of $\mathbf{v}_s$ and $\mathbf{c}_i$, this variant feeds their concatenation through a feed-forward network:
\begin{align}
f(s,i)=\mathbf{w}_a^{\top} \tanh (\mathbf{W}_a[\mathbf{v}_s;\mathbf{c}_i])\text{.}
\end{align}
Here $\mathbf{w}_a\in \mathbb{R}^{d'}$ and $\mathbf{W}_a\in \mathbb{R}^{d'\times 2d}$ are parameters of the feed-forward network.
By involving additional parameters $\mathbf{w}_a$ and $\mathbf{W}_a$, we expect the \textsc{concat} variant to be more capable than \textsc{dot}.

\paratitle{\textsc{trans-dot}}\quad In \textsc{dot}, we  multiply the seed and word vectors $\mathbf{v}_s$ and $\mathbf{c}_i$.
Note that the context word vectors $\mathbf{c}_i$ need to both interact with the seed vector $\mathbf{v}_s$ and constitute the context representation $\mathbf{v}_C$.
To distinguish between the two potentially different roles, we additionally consider the following \textsc{trans-dot} scoring function:
\begin{align}\label{eq:trans-dot}
f(s,i)=\mathbf{v}_s^{\top} \tanh (\mathbf{W}_a \mathbf{c}_i)\text{.}
\end{align}
Here, we use a fully connected layer with parameters $\mathbf{W}_a\in \mathbb{R}^{d\times d}$ to transform $\mathbf{c}_i$ before taking a dot product with $\mathbf{v}_s$.
Compared with \textsc{dot}, the \textsc{trans-dot} scoring function only introduce a medium-sized parameter space, which is smaller than that of \textsc{concat}.

In order to apply seed-aware attention to our network structure, we  use the respective scoring functions $f(s,i)$ to replace $f(i)$ in Eq.~\ref{eq:softmax}.
The resulted attention weights $\alpha_{s,i}$ are fed to Eq.~\ref{eq:attn_combine}, and make the context vector $\mathbf{v}_C$ seed-aware.

\section{Experimental Settings}\label{sec:settings}

\subsection{Dataset Processing}

\begin{table}\small
  \centering
  \begin{tabular}{lr}
  \toprule
  \textbf{Item}&\textbf{Count}\\
  \midrule
  Number of sentences&1,847,717\\
  Number of training sentences $|\mathcal{T}|$&1,478,173\\
  Number of testing sentences&369,544\\
  Average number of context words $|C|$&31.39\\
  Average number of hyponym terms $|T|$&3.46\\
  Number of unique terms&182,167\\
  Number of unique terms on training set $|L|$&180,684\\
  Vocabulary size of all contexts&941,603\\
  Vocabulary size of all training contexts&\multirow{2}{*}{119,270}\\(discarding words with freq $<5$) &\\
  \bottomrule
  \end{tabular}
  \caption{Summary of the derived dataset.}\label{tab:dataset}
\end{table}

We earlier briefed that CASE exploits sentences with Hearst patterns for training and evaluation.
For this reason, large-scale natural annotations can be easily obtained without manual effort.

Specifically, we employ an existing web-scale dataset, WebIsA\footnote{\url{http://webdatacommons.org/isadb/}}~\cite{seitner2016large}, to derive large-scale annotated sentences.
This dataset has 400 million hypernymy relations, extracted from 2.1 billion web pages.
For each hyponym-hypernym pair, the dataset provides IDs of source sentences and matched patterns where the pair occurs.
For example, a sentence ``\emph{Young barley grass is high in vitamin, antioxidant, enzyme, mineral, amino acid, chlorophyll and other phyto-nutrients.}'' leaves its ID and pattern ``\dots and other \dots'' in the lists of hypernymy pairs ``vitamin $\rightarrow$ phyto-nutrient'', ``antioxidant $\rightarrow$ phyto-nutrient'', \etc
Precisions of all patterns are also summarized as global information.
We use the information to decompose the sentence,
obtaining the example in the Dataset and Formal Task Definition section.
Specifically, we follow the below steps.
\begin{enumerate}
  \item We convert all words to lowercase and lemmatize them.
  \item We then filter the dataset with the pattern precision information, due to the noisy web pages and the error-prone hypernymy extraction procedure.
That is, we identify and keep \emph{high-quality sentences} where a hypernym is extracted with at least three hyponyms by a pattern with precision $\ge 0.5$.
  \item We regard hyponym terms appearing in at least ten high-quality sentences as \emph{high-quality terms}. We select high-quality sentences with at least three high-quality terms in the final dataset.
\end{enumerate}

Finally, our dataset contains 1,847,717 naturally labeled sentences, involving over 180k hyponym terms.
From them, we sample 20\% of sentences to form the test set, and use the remainder for training.
Table~\ref{tab:dataset} summarizes our dataset.


\subsection{Baseline Approaches}\label{sec:baseline}

Since no previous study addresses the exact CASE task, we evaluate our models against the solutions proposed for the most similar task, \ie lexical substitution.
Specifically, we compare with \citenoun{melamud2015simple}'s unsupervised method and one of its variants. We also evaluate a supervised method by~\citenoun{roller2016pic}.

\noindent \textsc{\textbf{L}exical \textbf{S}ubstitution} (\textsc{\textbf{LS}}). Word embedding models such as \citenoun{mikolov2013distributed} compute two types of word vectors, \ie IN and OUT.
\citenoun{melamud2015simple}'s analysis suggests that the IN-IN similarity favors synonyms or words with similar functions, while the IN-OUT similarity characterizes word compatibility or co-occurrence.
By promoting terms $t$ having the same meaning with the seed $s$ and good compatibility with the context $C$, they score a term $t$ by
\begin{equation}\label{eq:LS}
\textsc{LS}=\lambda_1 cos(\mathbf{s}^{I},\mathbf{t}^{I})+\frac{1-\lambda_1}{|C|}\sum_{c\in C} cos(\mathbf{c}^{I},\mathbf{t}^{O})
\end{equation}
Here the superscripts $I$ and $O$ stands for IN and OUT, respectively.
We train word vectors on all sentences in $\mathcal{T}$, and use averaged vectors to represent multi-word terms.
We follow the original paper and set $\lambda_1=0.5$.

\noindent \textsc{\textbf{LS} with Term \textbf{Co}-occurrence} (\textsc{\textbf{LSCo}}). Considering that expansion terms are not simply synonyms of seeds, and tend to co-occur with seeds (in Hearst patterns), we also study a modified version of Eq.~\ref{eq:LS}:
\begin{equation}\label{eq:LSCo}
\textsc{LSCo}=\lambda_2 \textsc{LS}+(1-\lambda_2)cos(\mathbf{s}^{I},\mathbf{t}^{O})
\end{equation}
We tune $\lambda_2$ and adopt the best-effort results.

\noindent \textsc{\textbf{P}robability \textbf{I}n \textbf{C}ontext} (\textsc{\textbf{PIC}})~\cite{roller2016pic}.
Different from the second term of Eq.~\ref{eq:LS}, \textsc{PIC} models the context compatibility by introducing a parameterized linear transformation on $\mathbf{c}^{I}$.
Therefore, it needs data to train the additional parameters and is inherently supervised.

\subsection{Parameters and Evaluation Metrics}

We trim or pad all contexts to length 100, and treat words occurring less than 5 times as OOVs.
Word vectors are pre-trained with \texttt{cbow}~\cite{mikolov2013distributed}.
Their dimensions $d$ as well as encoded contexts' and seeds' are set to 100.
The intermediate dimension $d'$ of attention-related network is set to 10.
Each batch is of size 128 with 1,000 negative samples to compose the sampled candidates.
We iterate for 10 epoches with the Adam optimizer.
All other hyper-parameters are found to work well by default and not tuned.

\begin{figure}
  \centering
  \includegraphics[width=0.69\columnwidth]{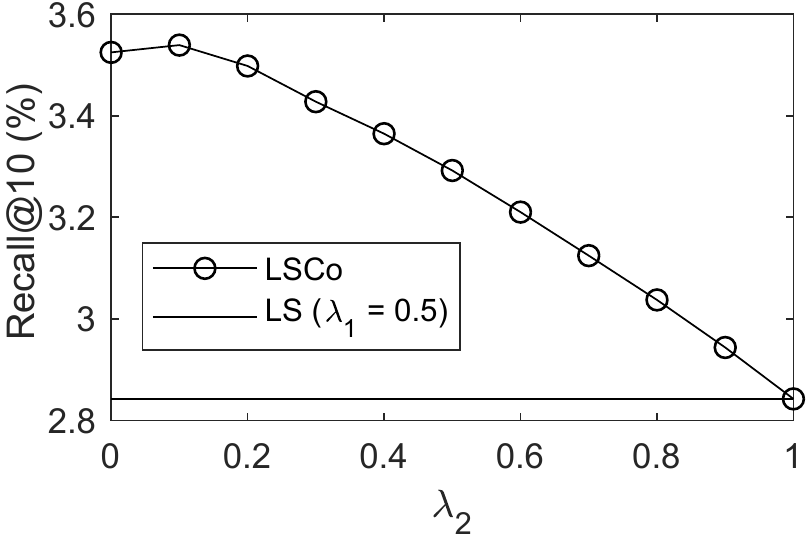}
  \caption{Tuning $\lambda_2$ in the \textsc{LSCo} baseline.}\label{fig:LS_LSCo}
\end{figure}

\begin{table}\small
\centering
  \setlength\tabcolsep{5pt}
  \begin{tabular}{l|rrrr}
  \toprule
  \textbf{Model}&\textbf{Recall}&\textbf{MAP}&\textbf{MRR}&\textbf{nDCG}\\
  \midrule
\textsc{LS}~\cite{melamud2015simple}&2.84&1.80&1.79&1.66\\
\textsc{LSCo} ($\lambda_2=0.1$)&3.54&2.65&2.69&2.23\\
\textsc{PIC}~\cite{roller2016pic}&19.62&17.78&18.84&14.71\\
\midrule
\textsc{CASE} (ours, with \textsc{NBoW})&\textbf{23.42}&\textbf{21.30}&\textbf{22.71}&\textbf{17.80}\\
\bottomrule
  \end{tabular}
  \caption{Comparison with LS baselines (top-10 results).}\label{tab:ls_results}
\end{table}

For all approaches, we uniformly rank all $t\in L$ according to the corresponding probability.
We concentrate on top-10 results.
Note that, due to the nature of natural language, ground-truth term lists may not be exhaustive.
This is an intrinsic limitation of the original dataset, and our processed dataset is probably the best we can access.
To this end, we use Recall as the main metric and do not involve Precision.
We also report MAP, MRR, and nDCG for reference.

\section{Experimental Results}\label{sec:results}

In this section,
we aim to experimentally answer the following questions:
\textbf{1)} Are lexical substitution solutions applicable to CASE?
\textbf{2)} Do contexts have impact on semantic expansion?
\textbf{3)} Is seed-aware attention superior as expected?
\textbf{4)} Do additional hypernyms make the experiments biased?


\subsection{Comparison with LS Baselines}

When introducing \citenoun{melamud2015simple}'s lexical substitution baseline, we mention that expansion terms should co-occur with, rather than be synonyms of, the seed term.
In Figure~\ref{fig:LS_LSCo}, we compare the Recall@10 scores of baselines \textsc{LS} and \textsc{LSCo}, \emph{w.r.t.}\ different $\lambda_2$.
Note that \textsc{LSCo} degenerates to \textsc{LS} when $\lambda_2=1$, so their lines overlap at this point.
The figure demonstrates that, when $\lambda_2<1$, the \textsc{LSCo} baseline outperforms \textsc{LS}, and achieves optimum when $\lambda_2=0.1$.

In Table~\ref{tab:ls_results}, we report the top-10 metrics of all three lexical substitution baselines, as well as those of our approach with the preliminary \textsc{NBoW} encoder.
By additionally advocating co-occurrence between $s$ and $t$, \textsc{LSCo} outperforms \textsc{LS} on all metrics.
However, it is remarkably inferior due to its unsupervised nature.

By parameterizing the context compatibility in \textsc{LS}, \textsc{PIC} achieves reasonably better results.
However,  \textsc{PIC} only models the similarity of seed and expansion terms through non-parameterized IN-IN similarity like the first term in Eq.~\ref{eq:LS}.
This may be inadequate, with reasons similar to the inferiority of \textsc{LS} to \textsc{LSCo}.
In our solution, the embedding-initialized parameters allow our seed encoder and prediction layer to capture type-based similarity beyond IN-IN and IN-OUT through training.
With the simplest \textsc{NBoW} encoder, the joint training of the two components helps our approach outperform \textsc{PIC} by a large margin.

\begin{table}\small
\centering
  \setlength\tabcolsep{7pt}
  \begin{tabular}{l|rrrr}
  \toprule
  \textbf{Context Encoder}&\textbf{Recall}&\textbf{MAP}&\textbf{MRR}&\textbf{nDCG}\\
  \midrule
No Encoder&15.81&14.02&14.85&11.62\\
\midrule
\underline{\textbf{RNN-Based}}&&&&\\
\textsc{RNN-vanilla}&17.51&16.17&17.08&13.26\\
\textsc{GRU}&18.99&17.28&18.31&14.24\\
\textsc{LSTM}&19.02&17.40&18.43&14.31\\
\textsc{BiLSTM}&14.59&13.45&14.22&10.96\\
\midrule
\textsc{CNN}&20.97&19.40&20.61&15.94\\
\midrule
\underline{\textbf{Placeholder-Aware}}&&&&\\
\textsc{CNN+PF}&20.88&19.04&20.20&15.70\\
\textsc{context2vec}&20.21&18.53&19.66&15.29\\
\midrule
\textsc{NBoW}&\textbf{23.42}&\textbf{21.30}&\textbf{22.71}&\textbf{17.80}\\
\bottomrule
  \end{tabular}
  \caption{Performance of context encoders (top-10 results).}\label{tab:context_results}
\end{table}



\begin{table*}\small
\centering
  \setlength\tabcolsep{7.5pt}
  \begin{tabular}{l|rrr|rrr|rrr|rrr}
  \toprule
  \multirow{2}{*}{\textbf{Model}}&\multicolumn{3}{c|}{\textbf{Recall}}&\multicolumn{3}{c|}{\textbf{MAP}}&\multicolumn{3}{c|}{\textbf{MRR}}&\multicolumn{3}{c}{\textbf{nDCG}}\\
  \cline{2-13}
  &@5&@10&@20&@5&@10&@20&@5&@10&@20&@5&@10&@20\\
  \midrule
\textsc{LSTM}&13.08&19.02&26.19&16.80&17.40&17.14&17.15&18.43&19.13&11.88&14.31&16.73\\
\quad\textsc{+attn}&13.73&19.85&27.15&17.72&18.29&17.96&18.11&19.41&20.11&12.55&15.04&17.51\\
\midrule
\textsc{NBoW}&16.32&23.42&31.64&20.78&21.30&20.79&21.29&22.71&23.45&14.91&17.80&20.58\\
\quad\textsc{+attn}&16.69&23.88&32.24&21.36&21.87&21.30&21.89&23.32&24.06&15.29&18.22&21.06\\
\quad\textsc{+dot}&15.54&22.13&29.89&20.12&20.60&20.12&20.61&21.95&22.66&14.36&17.03&19.66\\
\quad\textsc{+concat}&16.85&24.12&32.53&21.57&22.04&21.47&22.10&23.54&24.28&15.46&18.41&21.27\\
\quad\textsc{+trans-dot}&\textbf{17.20}&\textbf{24.51}&\textbf{33.01}&\textbf{21.97}&\textbf{22.41}&\textbf{21.80}&\textbf{22.53}&\textbf{23.96}&\textbf{24.70}&\textbf{15.80}&\textbf{18.77}&\textbf{21.65}\\
  \bottomrule
  \end{tabular}
  \caption{Performance of different scoring functions in attention. \textsc{trans-dot} is significantly better at $p<0.01$.}\label{tab:attn_results}
\end{table*}

\subsection{Comparison of Context Encoders}

The Introduction section mentioned that set expansion is similar to CASE without context.
We find that one seed is usually sufficient to retrieve terms of the same type.
The result thus heavily depend on the context to pick the right terms out of many others with the same type.
Table~\ref{tab:context_results} reflects this by the inferior results of the ``No Encoder'' setting, where contexts are removed in both training and testing.

Although contexts are important, complex encoders do not necessarily lead to better results.
In Table~\ref{tab:context_results}, encoders at lower semantic levels, \ie \textsc{NBoW} at the word level and \textsc{CNN} at the phrase level, are the most effective.
Among them, the simpler \textsc{NBoW} achieves better scores.
Moreover, RNN-based ones are not very competitive,
with the best \textsc{LSTM} variation poorer than \textsc{CNN}.
This may be due to that RNNs are only effective where predictions are sensitive to word orders, \eg in POS tagging and dependency parsing.
Finally, being placeholder-aware, the \textsc{context2vec} encoder performs better than its \textsc{LSTM} counterpart.
However, \textsc{CNN} with positional embedding, the stronger placeholder-aware encoder, is inferior to its \text{CNN} counterpart.
This indicates that CASE is inherently different from tasks like relation classification and aspect/targeted sentiment analysis, which rely on relative position between the placeholder and some key words.

\begin{table}\small
\setlength\tabcolsep{6pt}
\centering
\begin{tabular}{cc|cc}
\toprule
\multicolumn{2}{c|}{\textbf{With Hypernym}}&\multicolumn{2}{c}{\textbf{Without Hypernym}}\\
\midrule
\textsc{NBoW}&\textsc{+trans-dot}&\textsc{NBoW}&\textsc{+trans-dot}\\
\midrule
protein&\textbf{mineral}&protein&\textbf{mineral}\\
sugar&sugar&calcium&etc\\
\textbf{vitamin}&protein&salt&sugar\\
\textbf{mineral}&\textbf{vitamin}&sugar&\textbf{vitamin}\\
carbohydrate&b vitamin&\textbf{vitamin}&protein\\
herb&\textbf{enzyme}&\textbf{enzyme}&herb\\
\textbf{enzyme}&amino acid&herb&carbohydrate\\
fat&herb&potassium&salt\\
fiber&\textbf{antioxidant}&\textbf{mineral}&fat\\
salt&salt&etc&vitamin c\\
\bottomrule
\end{tabular}
\caption{Case study on attention and hypernyms.}\label{tab:case}
\end{table}

Based on the above observations, we confirm that contexts have major impacts on CASE and deserve appropriate modeling.
However, complex encoders are inferior because CASE is insensitive to either word orders or seed term positions.
Modeling these signals leads to more unnecessary parameters to learn and brings in noises.

\subsection{Effectiveness of the Attention Mechanism}

In previous sections, we proposed two types of scoring functions to incorporate the attention mechanism in the context encoder.
In Table~\ref{tab:attn_results}, we denote the vanilla seed-oblivious attention by \textsc{attn}, and the three seed-aware functions by their names, respectively.
Due to the relatively small margin between the scores of different functions, we report the metrics for top-5 and 20 results in addition to top-10.
Although seed-aware attention is applicable to \textsc{LSTM}, we do not include the results since they do not outperform the corresponding combinations of \textsc{NBoW}.
The limited improvement may be due to the low potential of the base \textsc{LSTM} encoder.

Table~\ref{tab:attn_results} shows that seed-oblivious attention can improve both \textsc{LSTM} and \textsc{NBoW}.
Although seed-aware, the \textsc{dot} scoring function turns out to adversely affect the quality of expansion terms.
We speculate that the two different roles of context word vectors $\mathbf{c}$ render the simple dot function insufficient to characterize its interactions with $\mathbf{v}_s$.
The \textsc{concat} function, on the other hand, partially demonstrates  superiority of seed-aware attention with limited improvement over \textsc{attn}.
By slightly modifying \textsc{dot} with even fewer additional parameters than \textsc{concat}, \textsc{trans-dot} outperforms all competitors.
Further paired t-tests show that the superiority of \textsc{trans-dot} (as well as the most competitive runs in Tables~\ref{tab:ls_results} and \ref{tab:context_results}) to all competitors is significant at $p<0.01$.
We attribute the statistical significance to the huge size of our testing set, \ie 369,544 sentences.

To illustrate the impact of \textsc{trans-dot}, we show expansion terms of ``amino acid'' for the example in the Dataset and Formal Task Definition section, in the first two columns of Table~\ref{tab:case}. Observe that \textsc{trans-dot}-based attention helps promote the ground truth terms (in bold) in the ranking.
It also removes nutrition ``fat'' from the top results, which is irrelevant to barley grass.

\subsection{Impacts of Hypernyms}

\begin{table}\small
\setlength\tabcolsep{6pt}
\begin{tabular}{l|rrrr}
\toprule
\textbf{Model} (w/o Hypernym)&\textbf{Recall}&\textbf{MAP}&\textbf{MRR}&\textbf{nDCG}\\
\midrule
\textsc{NBoW}&22.64&20.68&22.03&17.22\\
\quad\textsc{+trans-dot}&23.41&21.52&22.98&17.94\\
\bottomrule
\end{tabular}
\caption{Scores after removing hypernyms (top-10 results).}\label{tab:no_hypernym}
\end{table}

The contexts from WebIsA always contain hypernyms, \eg ``phyto-nutrients'' in the example of the Dataset and Formal Task Definition section.
However, practical scenarios may involve sentences without hypernyms as in Figure~\ref{fig:example}.
To study the potential impact, we remove all hypernyms in contexts, retrain and test \textsc{NBoW} with or without \textsc{trans-dot}.
The last two columns in Table~\ref{tab:case} show the results of our running example without the suffix ``and other phyto-nutrients''.
It is observed that removing hypernyms causes some non-nutrient or noisy terms (\eg ``salt'' and ``etc'') to rise.
Table~\ref{tab:no_hypernym} reports the overall scores for top-10 results.
Compared with the corresponding results in Table~\ref{tab:attn_results},  all scores slightly decrease by around one point.
This comparison suggests that, trained with sufficient term co-occurrences, our model is able to find terms of the same types, without the help of hypernyms in most cases.
To conclude, the hypernym bias introduced by the data harvesting approach has very small impacts on the practical use of our solution.

\section{Related Work}


\paratitle{Lexical Substitution}
This task has been investigated for over a decade~\cite{mccarthy2007semeval}.
It differs from CASE in that the substitutes are required to preserve the \emph{same} meaning with the original word.
Previous solutions follow two stages, \ie \emph{candidate generation} and \emph{candidate ranking}.
Synonym candidates are generally generated from external dictionaries or by pooling the testing data.
The ranking stage then boils down to estimating the compatibility between candidates and the context.

\citenoun{giuliano2007fbk} rely on n-grams to model candidates' compatibility.
\citenoun{erk2008structured} argues that syntactic relations in contexts are crucial, \eg ``a horse draws something'' and ``someone draws a horse''.
In~\citenoun{melamud2015simple}, word vectors~\cite{mikolov2013distributed} are applied to score candidates' similarity with the original word and their context compatibility.
Their method is nearly state-of-the-art, yet remains relatively simple.
Besides unsupervised approaches, supervised methods~\cite{szarvas2013supervised,szarvas2013learning,roller2016pic} prove superior at the cost of requiring more annotations.
We have experimentally compared with representative ones from both categories.

\paratitle{Set Expansion} This task aims to expand a couple of seeds to more terms in the underlying semantic class.
Most existing approaches involve bootstrapping on a large corpus of web pages~\cite{tong2008system,wang2007language,he2011seisa,chen2016long} or free text~\cite{shi2010corpus,shen2017setexpan,shi2014probabilistic,thelen2002bootstrapping}.
HTML-tag-based or lexical patterns covering a few seeds are extracted, which are then applied to the same corpus for new terms.
The process is iterated until certain stopping criterion is met.

Both this task and ours face the challenge of ambiguous terms, \eg ``apple''.
With multiple seeds, set expansion may rely on the other seeds, \eg ``samsung'' or ``orange'', for disambiguation.
However, since CASE accepts only one seed as input, it is essential to model the additional context to make up for the scarce information.
To this end, we resort to neural networks, where many off-the-shelf context modeling architectures are available.

\paratitle{Multi-Sense or Contextualized Word Representation} This technique deals with sense-mixing in traditional word representation.
Traditional word representations assign a single vector to each word.
They mix different senses of polysemous words, and block downstream tasks from exploiting the sense information.
\citenoun{reisinger2010multi} cluster the contexts of polysemous words and represent senses by the cluster centroids.
By sequentially carrying out context clustering, sense labeling, and representation learning, \citenoun{huang2012improving} obtain low-dimensional sense embeddings.
Non-parametric~\cite{neelakantan2014efficient} and probabilistic models with fewer parameters~\cite{tian2014probabilistic} are proposed later to accelerate training.

In multi-sense embedding, polysemous words get static embeddings for coarse-grained senses.
Some recent efforts explore dynamic embeddings that vary with the context.
\citenoun{melamud2015modeling} use context-aware substitutions of target words to obtain contextualized embeddings.
\citenoun{peters2018deep} employ multi-layered bi-directional language models on words in contexts.
Embeddings are obtained by aggregating different hidden layers with task-specific weights.
CASE separately models contexts and seed terms, because the model needs to generalize to unseen multi-word seeds.
For more studies, we refer readers to a survey~\cite{camacho2018word}.


\section{Conclusion}\label{sec:conclusion}

We define and address context-aware semantic expansion.
To the best of our knowledge, this is the first study on this task.
To facilitate training and evaluation without human annotations, we derive a large dataset with about 1.8 million naturally annotated sentences from WebIsA.
We propose a network structure, and study different alternatives of the context encoder.
Experiments show that solutions for lexical substitution are not competitive on CASE.
Comparisons on various context encoders indicate that, the simplest \textsc{NBoW} encoder achieves surprisingly good performance.
Based on \textsc{NBoW}, seed-aware attention, which models the interaction between seed and context words, further improves the performance.
The \textsc{trans-dot} scoring function finally shows its capability to focus on indicative words, and outperforms other seed-oblivious or -aware competitors.
In further analysis, we also confirm small impacts of a bias introduced when harvesting our data.

\bibliographystyle{aaai}
\bibliography{CASE}

\end{document}